%% file: 0_paper.tex
\documentclass[11pt,a4paper]{article}
\usepackage[hyperref]{stylefiles/emnlp-ijcnlp-2019}
\usepackage{times}
\usepackage{latexsym}

\usepackage{url}

\usepackage[T1]{fontenc}
\usepackage{mathtools}
\usepackage{multirow}
\usepackage{amsmath}
\usepackage{amssymb}
\usepackage{booktabs}
\usepackage{rotating}
\usepackage{enumitem}
\usepackage{todonotes}
\usepackage[linewidth=1pt]{mdframed}
\usepackage{url}
\usepackage{latexsym}

\aclfinalcopy 

\title{A Time Series Analysis of Emotional Loading in Central Bank Statements\vspace*{.3cm}}

\author{
Sven Buechel$^\clubsuit$ \hspace*{.1cm}
Simon Junker$^\diamondsuit$ \hspace*{.1cm}
Thore Schlaak$^\diamondsuit$ \hspace*{.1cm}
Claus Michelsen$^\diamondsuit$ \hspace*{.1cm}
Udo Hahn$^\clubsuit$ \vspace*{.3cm} \\
$^{\clubsuit}$ Jena University Language \& Information Engineering (JULIE) Lab\\
Friedrich-Schiller-Universit\"at Jena  \\
F\"urstengraben 27, D-07743 Jena, Germany \\
\url{https://julielab.de/}\vspace*{.3cm} \\
$^\diamondsuit$ German Institute for Economic Research (DIW) \\
Mohrenstra{\ss}e 58, D-10117 Berlin, Germany\\
\url{https://www.diw.de}
}
\date{}

\begin{document}
\maketitle
\begin{abstract}
  We examine the affective content of central bank press statements using emotion analysis. Our focus is on two major international players, the European Central Bank (ECB) and the US Federal Reserve Bank (Fed), covering a time span from 1998 through 2019. We reveal characteristic patterns in the emotional dimensions of valence, arousal, and dominance and find---despite the commonly established attitude that emotional wording in central bank communication should be avoided---a correlation between the state of the economy and particularly the dominance dimension in the press releases under scrutiny and, overall, an impact of the president in office.
  
\end{abstract}

\input{1_intro.tex}

\input{figs/example-statement.tex}
\input{figs/vadcube.tex}
\input{figs/vad-scatterplot.tex}
\input{figs/domseries.tex}
\input{2_data.tex}

\input{3_methods.tex}

\input{4_results.tex}
\input{5_discussion.tex}

\section*{Acknowledgments}
We would like to thank the anonymous reviewers for their detailed and constructive comments.

\bibliographystyle{stylefiles/acl_natbib}
\bibliography{literature-SB-econlp19}

\end{document}

%% file: 1_intro.tex
\section{Introduction}

Central Bank (henceforth, CB) communication has become increasingly important in the past 20 years for the world economy \cite{blinder2008central}. Until the mid-1990s, there was consensus that central bankers should remain more or less silent
and, if urged to make official statements, should try to hide their personal believes and assessments. This code of conduct changed fundamentally in recent years. Especially in times of unconventional monetary policy, central bankers are now trying to communicate proactively to economic agents, to give forward guidance and, thereby, try to increase the effectiveness of monetary policy \cite{Lucca09}. This has led to a fast growing economic literature about the content, type and timing of CB communications and the observable effects on the economy \citep[e.g.][]{ehrmann2007timing,ehrmann2007communication}.

CB communication and the reactions it causes are, in essence, verbally encoded---both in terms of official statements being issued as well as their assessment by other economic players and information gate-keepers (e.g., journalists, lobbyists).
Hence, more recent empirical work tries to incorporate NLP methods into economic analyses, e.g., using topic modeling \citep[e.g.][]{kawamura2019strategic} or information theory-based scores \cite{Lucca09}. However, these analyses are based on the assumption that statements by CBs are 
free from
emotions and contain factual information only.

But it is quite unlikely that even experienced communicators can fully hide their emotions in such a way that they cannot be traced by analytic means.
Hence, NLP methods might help reveal latent emotional loadings in CB communiqu\'{e}s.

Yet, if emotions can be identified, what is their added value for the interpretation of CB communication? In this paper, we intend to gather preliminary evidence that once emotional traces can be unlocked from CB communication, this additional information might help to better understand purely quantitative time series data signalling economic development congruent with emotional moves in CB press releases.


Regarding NLP, most previous work on emotion focused purely on \textit{polarity}, a rather simplified representation of the richness of human affective states in terms of positive--negative distinctions. For example, \citet{Nopp15} deal with sentiment analysis for exploring attitudes
and opinions about risk in textual disclosures issued by banks and derive sentiment scores that quantify uncertainty, negativity, and positivity in the analyzed documents (a collection
of more than 500 CEO letters and outlook
sections extracted from bank annual reports). The analysis of aggregated figures revealed strong and significant correlations between uncertainty
or negativity in textual disclosures and the
quantitative risk indicator's future evolution.

In contrast, a growing number of researchers start focusing on more complex and informative representations of affective states, often following distinct psychological research traditions \cite{Yu16,Wang16,Mohammad18acl,Buechel18naacl}.

Studies applying NLP methods to various other fields seem to benefit strongly from such additional information. For example, \citet{Kim17latech} examine the relationship between literary genres and emotional plot development finding that, in contrast to other, more predictive emotion categories, \textit{Joy} as a common emotional category is only moderately helpful for genre classification. More closely related to us, \citet{Bollen11} predict stock market prices based on Twitter data. They find evidence that more complex emotion measurements allow for more accurate predictions than polarity alone. The present study provides further evidence for this general observation focusing on the well-established emotional dimensions of \textit{Valence}, \textit{Arousal}, \textit{Dominance} (VAD) \cite{Bradley94} in CB statements.


To the best of our  knowledge, VAD measurements have neither been applied to analyzing verbal communication in the macro-economic field, in general, nor to CB communication, in particular. We show in this paper---based on the analysis of the press statements of the U.S. Federal Reserve (Fed) and the European Central Bank (ECB)---that CB communication is anything but free from emotions. We show that particularly the dimension of \textit{Dominance} is of high relevance and heavily depends on the state of the economy. Furthermore, communication also along the \textit{Valence} and \textit{Arousal} dimensions is largely affected by the individual CB presidents in office. Overall, this provides preliminary evidence that the presence of emotional loading in monetary policy communication, which is of high importance to central bankers, correlates with quantitative macro-economic indicators. Our findings provide promising avenues for further research, as the real effects of emotions on CB communication have largely been neglected in economic research.

%% file: figs/example-statement.tex
\begin{figure}[t!] 
    \begin{mdframed}
    \it \small
    Compared with the March 2019 ECB staff macroeconomic projections, the outlook for real GDP growth has been revised up by 0.1 percentage points for 2019 and has been revised down by 0.2 percentage points for 2020 and by 0.1 percentage points for 2021. The risks surrounding the euro area growth outlook remain tilted to the downside, on account of the prolonged presence of uncertainties, related to geopolitical factors, the rising threat of protectionism and vulnerabilities in emerging markets.
    \end{mdframed}
    \caption{\label{fig:statment}
    Excerpt of ECB statement from June 6, 2019. 
    }
\end{figure}

%% file: figs/vadcube.tex
\begin{figure}[b!]
    \centering
    \includegraphics[width=.5\textwidth]{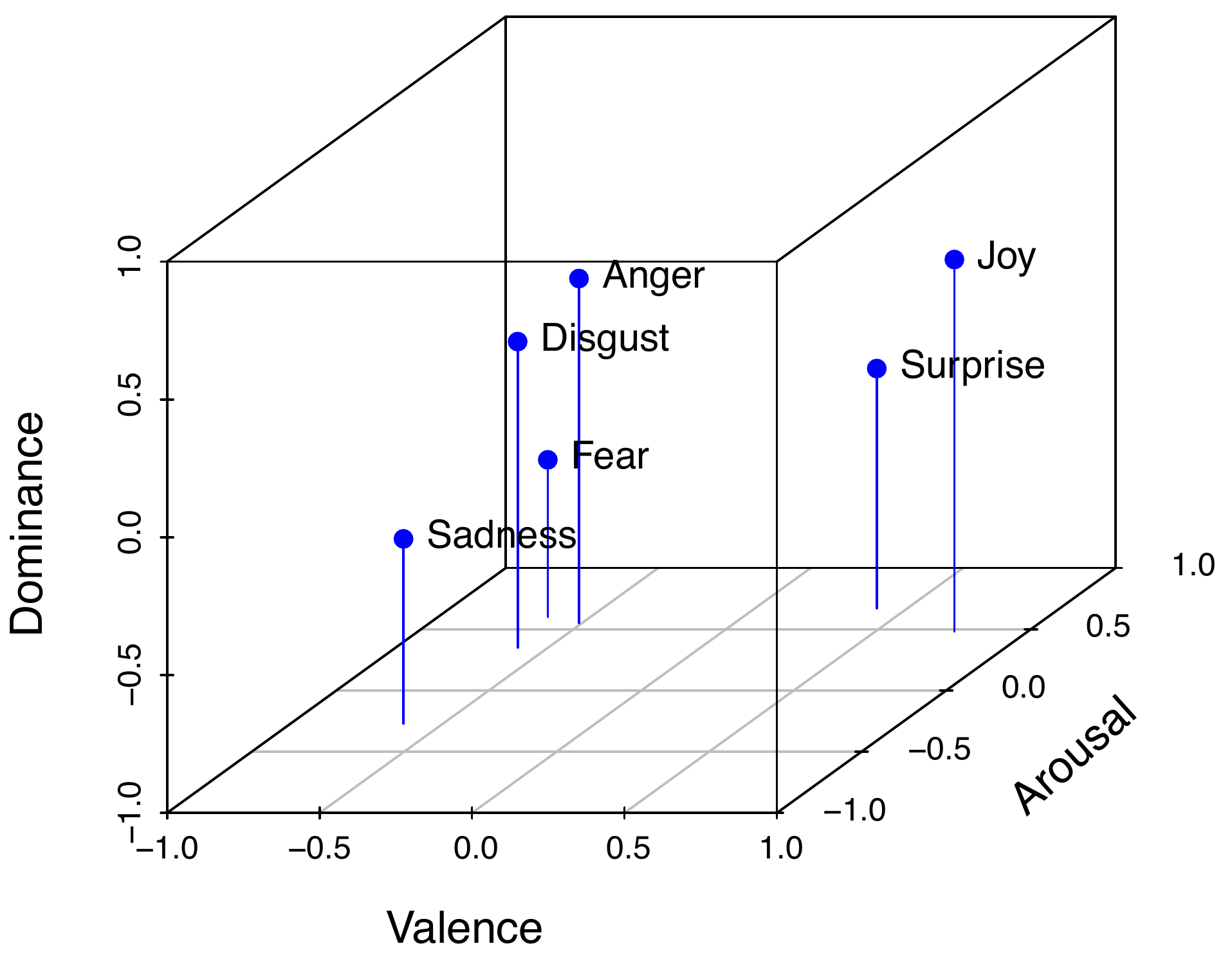}
    
    \caption{\label{fig:vadcube}
    Affective space spanned by the Valence-Arousal-Dominance (VAD) model, together with the position of six basic emotions. Adapted from \citet{Buechel16ecai}.} 
    \vspace*{-10pt}
\end{figure}

%% file: figs/vad-scatterplot.tex
\begin{figure*}
    \centering
    \includegraphics[width=.8\textwidth]{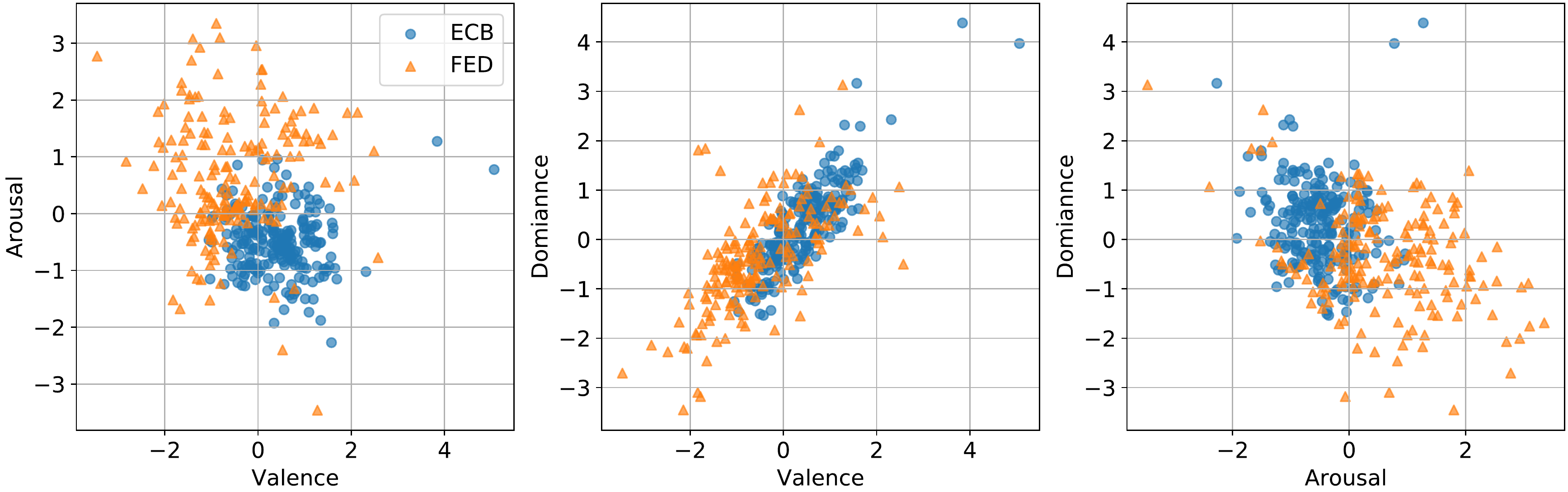}
    \caption{\label{fig:scatterplot}
    Scatterplot array of bivariate emotion distribution in ECB statements (blue circles) vs. Fed statements (orange triangles) with respect to\textit{ Valence} and \textit{Arousal} (left), \textit{Valence} and \textit{Dominance} (center), and \textit{Arousal} and \textit{Dominance} (right). VAD scores are centered and scaled.}
\end{figure*}

%% file: figs/domseries.tex
\begin{figure*}[t]
    \centering
    \includegraphics[width=.85\textwidth]{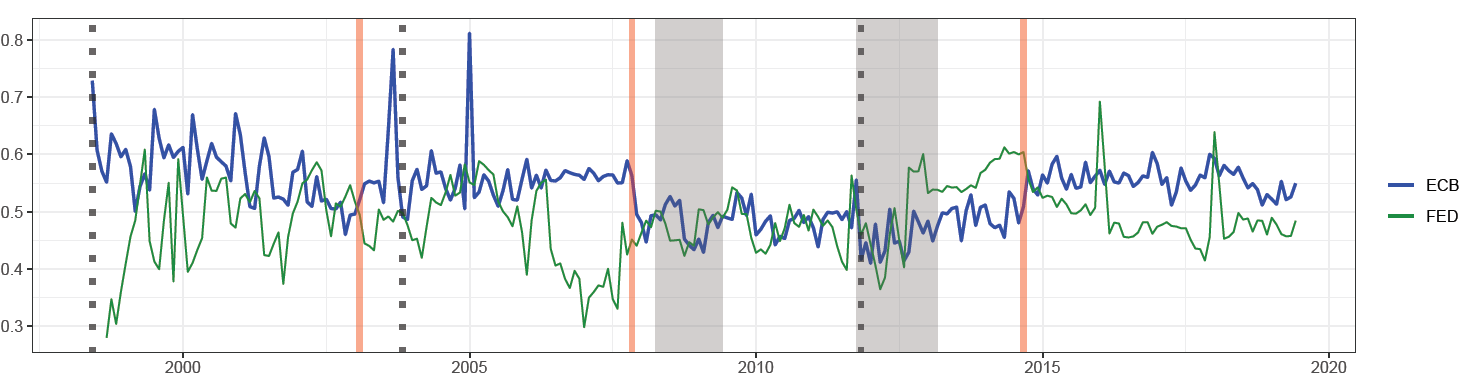}
    \caption{\label{fig:domseries}
    Dominance series for ECB (blue line) and Fed (green). Vertical dotted lines indicate beginning of ECB presidency (Duisenberg: 1998, Trichet: 2003, Draghi: 2011). Red vertical (solid) lines indicate break dates. Shaded areas highlight Euro area recession periods.}
\end{figure*}

%% file: 2_data.tex
\section{Data}
\label{sec:data}

We Web-scraped the policy statements issued by the ECB and the Fed from their Web pages, starting with the first communiqu\'{e} by the ECB when it formally replaced the European Monetary Institute in June 1998. The most recent documents for both ECB and Fed have been issued in June 2019. These statements contain an assessment of the economic situation by the CB, its policy decisions and the main arguments underlying them. Altogether we assembled 230 documents from the ECB and 181 from the Fed that contain on average 1583 and 417 tokens, respectively. 
We illustrate the particular style of these documents with an excerpt in Figure \ref{fig:statment}.

%% file: 3_methods.tex
\section{Methods}

Measuring the emotional content of natural language utterances has become a particularly rich area of research.  
The choice of an adequate emotion representation format, i.e., the mathematical domain of the label space and its interpretation in terms of psychological theory, has become a crucial aspect of computational emotion analysis (in contrast to work focusing only on polarity) \cite{Buechel17eacl}. 

The majority of prior work follows the so-called \textit{discrete} approach to emotion representation where a small set of universal \textit{basic emotions} \citep{Ekman92}, such as \textit{Joy}, \textit{Anger}, and \textit{Sadness}, is stipulated. Equally popular in psychology though is the \textit{dimensional} approach which represents emotions as real-valued vectors, having components such as \textit{\textbf{V}alence} (pleasure vs.\ displeasure), \textit{\textbf{A}rousal} (calmness vs.\ excitement) and \textit{\textbf{D}ominance} (being controlled by vs.\ having control over a social situation; \citealt{Bradley94}). Figure \ref{fig:vadcube} provides an illustration of these dimensions relative to common emotional categories. In this work, we employ the VAD format because of its greater flexibility, following our previous work \cite{Buechel-wassa16,Handschke18}.

Given the relatively high average token number per document in our corpus, we adopted a comparably simple lexicon-based approach which models document emotion based on word frequency combined with empirical measurements of lexicalized word emotions.\footnote{Although less reliable for individual sentences than neural methods, lexicon-based methods still perform well on longer documents since the larger amount of word material improves predictions based on word frequency statistics \citep{Sap14emnlp}.
} 
Such \textit{emotion lexicons} have a long tradition in psychology \citep{stone_general_1966} and are nowadays available for various emotion formats and many different language \citep{Buechel18coling}. Roughly speaking, their creation follows a questionnaire study-like design. For English VAD scores, the lexicon by \citet{warriner2013} is a common choice due its large coverage (14k lexical units) which we adopt as well.

For the computation of document-level VAD scores, we rely on the open-source tool \textsc{JEmAS} \newcite{Buechel16ecai}).\footnote{\url{https://github.com/JULIELab/JEmAS}} 
It estimates the emotion value of a document $d$, $\bar{e}(d)$, as weighted average of the empirical emotion values of the words in $d$, $e(w)$:
\begin{equation}
\label{eq:jemas}
\bar{e}(d) := \frac{\sum_{w \in d} \lambda(w,d) \times e(w)}{\sum_{w \in d} \lambda(w,d)}                       
\end{equation}
where $e(w)$ is defined as the vector representing the neutral emotion, if $w$ is not covered by the lexicon, and $\lambda$ denotes some term weighting function. Here, we use absolute term frequency.

For our subsequent time series study, we process the CB corpus (see Section \ref{sec:data}) using \textsc{JEmAS}. The result is one three-dimensional VAD value per document. As both an exploratory analysis and sanity check, we center and scale the resulting data and visualize them as scatterplot array (see Figure \ref{fig:scatterplot}). ECB statements are higher in \textit{Valence} and \textit{Dominance} but lower in \textit{Arousal} than Fed statements (in all cases $p < .001$; Mann--Whitney U test). As often observed \cite{warriner2013}, \textit{Valence} and \textit{Dominance} have a strong linear correlation ($r=.758$).

Since neither the ECB nor the Fed hold monthly meetings, the corresponding time series of the emotion scores have missing values across the sample at a monthly frequency. A standard procedure to deal with this is linear interpolation. This appears appropriate for the ECB, since its meetings always took place at a frequent and regular pace, resulting in only 11\% missing data points on a monthly basis. The Fed, however, successively increased the number of statements following their meetings. Initially, they only communicated after a policy change, but later decided to do so after each meeting. There are eight regular meetings per year plus additional sessions as required. This results in fewer data points than for the ECB, with roughly a third of data points missing.
   
\input{tabs/correlations.tex} 

In order to avoid artifacts due to the interpolation procedure, we alternatively apply the method of \citet{RePEc:zbw:bubdp1:5097}. They use the correlation between a series affected by missing values and another, complete time series to interpolate the missing data points. The linearly interpolated series of the emotion scores are highly correlated with a broad set of economic data in their respective geographic area (see Table \ref{tab:correlations}): we compiled data sets for the Euro area and the US covering a measure of the change in the real economy (approximated by industrial production), inflation, unemployment and interest rates. These are the main economic variables in most small-scale economic models. We add business and consumer survey data to incorporate forward looking elements and retail sales to complement the industry data with service sector-based information.

As one may expect, the \textit{Valence} measurements are correlated with all activity measures, at least for the ECB. This may be due to the description of the current state of the economy inherent in the statements, which necessarily apply words with a positive or negative connotation based on the business cycle phase. Interestingly, this does not hold for the Fed, and, moreover, the \textit{Arousal} and \textit{Dominance} scores are also highly correlated with economic variables, particularly with inflation and unemployment---the CBs' main (direct or intermediate) target variables. 

The Schumacher/Breitung procedure generates VAD-time series which, for the ECB, look virtually unchanged in comparison with the interpolated series, while the corresponding Fed scores are more volatile. For the latter reason, we stick to linear interpolation, while the results from the ECB case confirm that this method does not induce too much bias. Finally, to avoid interpolation altogether, we check whether the results persist under a qualitative perspective, if we repeat the following analysis on series aggregated to quarterly frequency.

%% file: tabs/correlations.tex
\hspace{-15pt}
\begin{table}[b!]
\centering \small
\begin{tabular}{lccccc}\hline
                                      &   & production & inflation & unemploy.  & services \\ \cline{3-6}
\multirow{3}{*}{ \rotatebox{90}{ECB}} & V &     ~0.32 &      ~0.19 &    -0.45 &  ~0.42 \\
                                      & A &    -0.11 &      ~0.24 &    -0.34 & -0.23 \\
                                      & D &     ~0.24 &     -0.12 &    -0.32 &  ~0.53 \\ \hline
    \multicolumn{5}{c}{ } \\[-0.3cm]
\multirow{3}{*}{ \rotatebox{90}{Fed}} & V &     ~0.04 &     -0.03 &     ~0.00 &  ~0.07 \\
                                      & A &     ~0.05 &      ~0.52 &    -0.56 &  ~0.12 \\
                                      & D &    -0.03 &     -0.17 &     ~0.10 & -0.03 \\ \hline
\end{tabular}
\caption{\label{tab:correlations}
VAD scores and their correlation with a broad set of economic indicators (excerpt).}
\end{table}

%% file: 4_results.tex
\section{Results}

We perform standard break tests on the VAD scores; they are designed to detect endogenous changes in the underlying statistical process (which we model as auto-regressive, moving-average). Since Augmented-Dickey-Fuller tests indicate that these series are non-stationary, we use detrended data and find that the results also hold for the data in first differences. Focusing on the ECB, the break tests endogenously reveal three break points for each sentiment series. It has to be emphasized that the applied break tests return an endogenous break date without any restrictions by the researcher. Thus, a break date returned in proximity to a specific event makes it likely that the hypothesis of this event being causal for the break will not be rejected. A specific event study, however, is left for future research. Focusing on the ECB series, it turns out that, interestingly, independent tests for the three series reveal neighbouring break dates---either occurring around the change in presidency or key economic events (see Figure \ref{fig:domseries} for the dominance series). 

The first break, in 2003, is close for the \textit{Valence} and \textit{Arousal} series (in July and September, respectively) and somewhat earlier for the \textit{Dominance} series (in February). The second break is detected in winter 2008/09; again the points are close for \textit{Valence} and \textit{Arousal} (January '09 and September '08, respectively) and earlier for \textit{Dominance} (in November '07, just a month before the global crisis originated in the US). The third break appears unrelated between the VAD series: it occurs in October 2011 for the \textit{Valence} series, in February 2013 for the \textit{Arousal} series and in September 2014 for the \textit{Dominance} series.

This illustrates that the breakpoints, by and large, either coincide with major economic turning points, or the change in presidency of the respective CB: the first one, when Wim Duisenberg was followed by Jean-Claude Trichet in October 2003. The second break is close to the outbreak of the Great Recession, which is a common feature in most economic data due to the massive impact the global recession had on most variables. Hence, not only did the economy change drastically at that time, but also the emotions expressed by President Trichet became different. The \textit{Dominance} series, in particular, expresses this phenomenon: With the outbreak of the crisis, the corresponding emotion scores decreased markedly and remained low until the change in presidency in fall 2011. Since Mario Draghi became ECB President this score started to recover, as evidenced by the clear uptrend; the third break point also tracks this; it occurred when the \textit{Dominance} scores settled down on a new, higher level.

%% file: 5_discussion.tex
\section{Conclusion}

The findings of our analysis are threefold: We showed that central bankers, assumed to be among the most technically talking economic agents (for reasons of an assumed and/or desired communication efficiency), are prone to emotions which, in addition, are strongly influenced by the economic situation. The Great Recession also left its mark in the emotions of President Trichet who, according to emotion analysis coupled with standard econometric tools, switched to a markedly more submissive language. Interestingly, this attitude slowly recovered towards a more dominant stance once Mario Draghi took office. Thus, finally, our analysis shows that CB communication depends much on the person presiding it, albeit the shift to a different emotional stance, e.g. in \textit{Dominance}, fades in only gradually.

%% file: 0_paper.bbl
\begin{thebibliography}{23}
\expandafter\ifx\csname natexlab\endcsname\relax\def\natexlab#1{#1}\fi

\bibitem[{Blinder et~al.(2008)Blinder, Ehrmann, Fratzscher, De~Haan, and
  Jansen}]{blinder2008central}
Alan~S Blinder, Michael Ehrmann, Marcel Fratzscher, Jakob De~Haan, and
  David-Jan Jansen. 2008.
\newblock Central bank communication and monetary policy: a survey of theory
  and evidence.
\newblock \emph{Journal of Economic Literature}, 46(4):910--945.

\bibitem[{Bollen et~al.(2011)Bollen, Mao, and Zeng}]{Bollen11}
Johan Bollen, Huina Mao, and Xiaojun Zeng. 2011.
\newblock \textsc{Twitter} mood predicts the stock market.
\newblock \emph{Journal of Computational Science}, 2(1):1--8.

\bibitem[{Bradley and Lang(1994)}]{Bradley94}
Margaret~M. Bradley and Peter~J. Lang. 1994.
\newblock Measuring emotion: the {Self-Assessment Manikin} and the semantic
  differential.
\newblock \emph{Journal of Behavior Therapy and Experimental Psychiatry},
  25(1):49--59.

\bibitem[{Buechel and Hahn(2016)}]{Buechel16ecai}
Sven Buechel and Udo Hahn. 2016.
\newblock Emotion analysis as a regression problem: Dimensional models and
  their implications on emotion representation and metrical evaluation.
\newblock In \emph{ECAI 2016 --- Proceedings of the 22nd European Conference on
  Artificial Intelligence. Including Prestigious Applications of Artificial
  Intelligence (PAIS 2016). The Hague, The Netherlands, August 29 - September
  2, 2016}, number 285 in Frontiers in Artificial Intelligence and
  Applications, pages 1114--1122, Amsterdam, Berlin, Washington, D.C. IOS
  Press.

\bibitem[{Buechel and Hahn(2017)}]{Buechel17eacl}
Sven Buechel and Udo Hahn. 2017.
\newblock \textsc{EmoBank}: studying the impact of annotation perspective and
  representation format on dimensional emotion analysis.
\newblock In \emph{EACL 2017 --- Proceedings of the 15th Conference of the
  European Chapter of the Association for Computational Linguistics. Valencia,
  Spain, April 3-7, 2017}, volume 2: Short Papers, pages 578--585. Association
  for Computational Linguistics (ACL).

\bibitem[{Buechel and Hahn(2018{\natexlab{a}})}]{Buechel18coling}
Sven Buechel and Udo Hahn. 2018{\natexlab{a}}.
\newblock Emotion representation mapping for automatic lexicon construction
  (mostly) performs on human level.
\newblock In \emph{COLING 2018 --- Proceedings of the 27th International
  Conference on Computational Linguistics: Main Conference. Santa Fe, New
  Mexico, USA, August 20-26, 2018}, pages 2892--2904. International Committee
  on Computational Linguistics (ICCL).

\bibitem[{Buechel and Hahn(2018{\natexlab{b}})}]{Buechel18naacl}
Sven Buechel and Udo Hahn. 2018{\natexlab{b}}.
\newblock Word emotion induction for multiple languages as a deep multi-task
  learning problem.
\newblock In \emph{NAACL-HLT 2018 --- Proceedings of the 2018 Conference of the
  North American Chapter of the Association for Computational Linguistics:
  Human Language Technologies}, volume 1, long papers, pages 1907--1918, New
  Orleans, Louisiana, USA, June 1--6, 2018.

\bibitem[{Buechel et~al.(2016)Buechel, Hahn, Goldenstein, H\"{a}ndschke, and
  Walgenbach}]{Buechel-wassa16}
Sven Buechel, Udo Hahn, Jan Goldenstein, Sebastian G.~M. H\"{a}ndschke, and
  Peter Walgenbach. 2016.
\newblock Do enterprises have emotions?
\newblock In \emph{WASSA 2016 --- Proceedings of the 7th Workshop on
  Computational Approaches to Subjectivity, Sentiment and Social Media Analysis
  @ NAACL-HLT 2016. San Diego, California, USA, June 16, 2016}, pages 147--153.

\bibitem[{Ehrmann and
  Fratzscher(2007{\natexlab{a}})}]{ehrmann2007communication}
Michael Ehrmann and Marcel Fratzscher. 2007{\natexlab{a}}.
\newblock Communication by central bank committee members: different
  strategies, same effectiveness?
\newblock \emph{Journal of Money, Credit and Banking}, 39(2-3):509--541.

\bibitem[{Ehrmann and Fratzscher(2007{\natexlab{b}})}]{ehrmann2007timing}
Michael Ehrmann and Marcel Fratzscher. 2007{\natexlab{b}}.
\newblock The timing of central bank communication.
\newblock \emph{European Journal of Political Economy}, 23(1):124--145.

\bibitem[{Ekman(1992)}]{Ekman92}
Paul Ekman. 1992.
\newblock An argument for basic emotions.
\newblock \emph{Cognition \& Emotion}, 6(3-4):169--200.

\bibitem[{H\"{a}ndschke et~al.(2018)H\"{a}ndschke, Buechel, Goldenstein,
  Poschmann, Duan, Walgenbach, and Hahn}]{Handschke18}
Sebastian G.~M. H\"{a}ndschke, Sven Buechel, Jan Goldenstein, Philipp
  Poschmann, Tinghui Duan, Peter Walgenbach, and Udo Hahn. 2018.
\newblock A corpus of corporate annual and social responsibility reports: 280
  million tokens of balanced organizational writing.
\newblock In \emph{ECONLP 2018 --- Proceedings of the 1st Workshop on Economics
  and Natural Language Processing @ ACL 2018. Melbourne, Victoria, Australia,
  July 20, 2018}, pages 20--31, Stroudsburg/PA. Association for Computational
  Linguistics (ACL).

\bibitem[{Kawamura et~al.(2019)Kawamura, Kobashi, Shizume, and
  Ueda}]{kawamura2019strategic}
Kohei Kawamura, Yohei Kobashi, Masato Shizume, and Kozo Ueda. 2019.
\newblock Strategic central bank communication: discourse analysis of the {Bank
  of Japan}'s monthly report.
\newblock \emph{Journal of Economic Dynamics and Control}, 100:230--250.

\bibitem[{Kim et~al.(2017)Kim, Pad\'{o}, and Klinger}]{Kim17latech}
Evgeny Kim, Sebastian Pad\'{o}, and Roman Klinger. 2017.
\newblock Investigating the relationship between literary genres and emotional
  plot development.
\newblock In \emph{LaTeCH-CLfL 2017 --- Proceedings of the 1st Joint SIGHUM
  Workshop on Computational Linguistics for Cultural Heritage, Social Sciences,
  Humanities and Literature @ ACL 2017. Vancouver, British Columbia, Canada,
  August 4, 2017}, pages 17--26. Association for Computational Linguistics
  (ACL).

\bibitem[{Lucca and Trebbi(2009)}]{Lucca09}
David~O. Lucca and Francesco Trebbi. 2009.
\newblock Measuring central bank communication: an automated approach with
  application to {FOMC} statements.
\newblock Technical Report Working Paper No. 15367, National Bureau of Economic
  Research (NBER), Cambridge, MA, USA.

\bibitem[{Mohammad(2018)}]{Mohammad18acl}
Saif~M. Mohammad. 2018.
\newblock Obtaining reliable human ratings of valence, arousal, and dominance
  for 20,000 english words.
\newblock In \emph{ACL 2018 --- Proceedings of the 56th Annual Meeting of the
  Association for Computational Linguistics. Melbourne, Victoria, Australia,
  July 15-20, 2018}, volume 1: Long Papers, pages 174--184.

\bibitem[{Nopp and Hanbury(2015)}]{Nopp15}
Clemens Nopp and Allan Hanbury. 2015.
\newblock Detecting risks in the banking system by sentiment analysis.
\newblock In \emph{EMNLP 2015 --- Proceedings of the 2015 Conference on
  Empirical Methods in Natural Language Processing. Lisbon, Portugal, 17-21
  September 2015}, pages 591--600, Red Hook/NY. Association for Computational
  Linguistics (ACL), Curran Associates, Inc.

\bibitem[{Sap et~al.(2014)Sap, Park, Eichstaedt, Kern, Stillwell, Kosinski,
  Ungar, and Schwartz}]{Sap14emnlp}
Maarten Sap, Gregory~J. Park, Johannes~C. Eichstaedt, Margaret~L. Kern,
  David~J. Stillwell, Michal Kosinski, Lyle~H. Ungar, and Hansen~Andrew
  Schwartz. 2014.
\newblock Developing age and gender predictive lexica over social media.
\newblock In \emph{EMNLP 2014 --- Proceedings of the 2014 Conference on
  Empirical Methods in Natural Language Processing. Doha, Qatar, October 25-29,
  2014}, pages 1146--1151. Association for Computational Linguistics (ACL).

\bibitem[{Schumacher and Breitung(2006)}]{RePEc:zbw:bubdp1:5097}
Christian Schumacher and J\"org Breitung. 2006.
\newblock {Real-time forecasting of GDP based on a large factor model with
  monthly and quarterly data}.
\newblock Discussion Paper Series 1: Economic Studies 2006/33, Deutsche
  Bundesbank.

\bibitem[{Stone et~al.(1966)Stone, Dunphy, and Smith}]{stone_general_1966}
Philip~J Stone, Dexter~C Dunphy, and Marshall~S Smith. 1966.
\newblock \emph{The {General} {Inquirer}: {A} {Computer} {Approach} to
  {Content} {Analysis}.}
\newblock MIT Press.

\bibitem[{Wang et~al.(2016)Wang, Yu, Lai, and Zhang}]{Wang16}
Jin Wang, Liang-Chih Yu, K.~Robert Lai, and Xuejie Zhang. 2016.
\newblock Dimensional sentiment analysis using a regional {CNN-LSTM} model.
\newblock In \emph{ACL 2016 --- Proceedings of the 54th Annual Meeting of the
  Association for Computational Linguistics. Berlin, Germany, August 7-12,
  2016}, volume 2: Short Papers, pages 225--230, Stroudsburg, PA. Association
  for Computational Linguistics (ACL).

\bibitem[{Warriner et~al.(2013)Warriner, Kuperman, and
  Brysbaert}]{warriner2013}
Amy~Beth Warriner, Victor Kuperman, and Marc Brysbaert. 2013.
\newblock Norms of valence, arousal, and dominance for 13,915 {English} lemmas.
\newblock \emph{Behavior Research Methods}, 45(4):1191--1207.

\bibitem[{Yu et~al.(2016)Yu, Lee, Hao, Wang, He, Hu, Lai, and Zhang}]{Yu16}
Liang-Chih Yu, Lung-Hao Lee, Shuai Hao, Jin Wang, Yunchao He, Jun Hu, K.~Robert
  Lai, and Xuejie Zhang. 2016.
\newblock Building {Chinese} affective resources in valence-arousal dimensions.
\newblock In \emph{NAACL-HLT 2016 --- Proceedings of the 2016 Conference of the
  North American Chapter of the Association for Computational Linguistics:
  Human Language Technologies. San Diego, California, USA, June 12-17, 2016},
  pages 540--545, Stroudsburg/PA. Association for Computational Linguistics
  (ACL).

\end{thebibliography}
